\documentclass[runningheads]{llncs}
\usepackage[T1]{fontenc}
\usepackage{graphicx,verbatim}
\usepackage{algorithm}
\usepackage{algorithmic}
\usepackage{amsmath}
\usepackage{amssymb}
\usepackage{booktabs}
\usepackage{multirow}
\usepackage{array}

\begin{document}

\title{RefTr: Recurrent Refinement of Confluent Trajectories for 3D Vascular Tree Centerlines}
\titlerunning{RefTr}

\author{Roman Naeem, David Hagerman, Jennifer Alvén, Fredrik Kahl}
\authorrunning{R. Naeem et al.}
\institute{Chalmers University of Technology \\
    \email{nroman@chalmers.se}}
    
\maketitle              

\begin{abstract}
Tubular tree structures such as blood vessels and lung airways are central to many clinical tasks, including diagnosis, treatment planning, and surgical navigation. Accurate centerline extraction with correct topology is essential, as missing small branches can lead to incomplete assessments or overlooked abnormalities. We propose RefTr, a 3D image-to-graph framework that generates vascular centerlines via recurrent refinement of confluent trajectories. RefTr adopts a Transformer-based Producer–Refiner architecture in which the Producer predicts candidate trajectories and a shared Refiner iteratively refines them toward the target branches. The confluent trajectory representation enables wh-ole branch refinement while explicitly enforcing valid topology. This recurrent scheme improves precision and reduces decoder parameters by 2.4$\times$ compared to the state-of-the-art. We further introduce an efficient non-maximum suppression algorithm for spatial tree graphs to merge duplicate branches and extend evaluation metrics to be radius-aware for robust comparison. Experiments on multiple public datasets demonstrate stronger overall performance, faster inference, and substantially fewer parameters, highlighting the effectiveness of RefTr for 3D vascular tree analysis. The code is available at \url{https://github.com/RomStriker/RefTr}. 


\end{abstract}

\section{Introduction}

Tubular trees such as blood vessels and lung airways are essential for material transport in the human body, and abnormalities in these networks are linked to many diseases. Accurate, topologically correct centerline extraction is essential for diagnosis and treatment planning~\cite{li2022human,moccia2018blood}, surgical navigation~\cite{huang2011interactive}, hemodynamic analysis~\cite{miraucourt2017blood}, and vascular morphometry~\cite{choi2021ct,khan2018three}. Missing small branches can lead to incomplete assessments and clinical risk, making high recall crucial. Centerline graphs provide a compact, interpretable representation, easier to annotate than dense masks and amenable to enforcing valid tree topology.

Segmentation-based methods~\cite{tetteh2020deepvesselnet,wittmann2025vesselfm} require costly voxel-level labels, post-processing skeletonization, and lack end-to-end graph prediction. Graph-based methods~\cite{prabhakar2024vesselformer,shit2022relationformer} predict nodes and edges but do not guarantee valid tree topology and scale poorly. Sequential tracking methods~\cite{naeem2025trexplorer,naeem2024trexplorer} iteratively generate centerlines but suffer from low recall due to imbalanced bifurcation/termination classification and cannot correct early errors.

To address these limitations, we propose RefTr, a parameter-efficient Transformer-based~\cite{carion2020end,vaswani2017attention} 3D image-to-graph model that generates centerline trees via recurrent refinement of confluent trajectories. RefTr follows a Producer–Refiner architecture where the Producer predicts multiple trajectory candidates rooted at the input patch center and a shared Refiner recurrently refines them toward the target trajectories, enhancing precision and reducing decoder parameters by 2.4$\times$ compared to the state-of-the-art Trexplorer Super~\cite{naeem2025trexplorer}. The confluent trajectory representation allows refinement of complete trajectories while explicitly encoding tree topology through predicted divergence and end positions, as illustrated in Figure~\ref{fig:arch}a.

RefTr maximizes recall and improves precision by predicting multiple candidates per target via many-to-one matching, creating an ensembling effect. This intentional overprediction introduces duplicate trajectories, which are suppressed at the patch level using predicted divergence positions between trajectory pairs, since duplicates do not diverge. Remaining duplicates at the global tree level are removed using Tree Non-Maximum Suppression (TNMS), a novel, efficient algorithm that merges overlapping branches in spatial tree graphs while preserving topology. For robust comparison, we extend the evaluation framework of Trexplorer Super~\cite{naeem2025trexplorer} with radius-aware thresholds that scale with vessel size, and report the average across multiple thresholds. Our concurrent submission~\cite{anon2026} focuses on EVAR clinical deployment, whereas this work addresses the methodological problem of general vascular tree reconstruction.

Our contributions include: (1) RefTr, a parameter-efficient 3D image-to-graph framework that predicts confluent trajectories and achieves strong overall performance, faster inference, and substantially fewer parameters compared to the state-of-the-art; (2) Tree Non-Maximum Suppression, an efficient algorithm for removing duplicate branches while preserving tree topology; and (3) radius-aware evaluation metrics for robust comparison across datasets with varying vessel sizes.

\section{Method}

\begin{figure}[!t]
\centering
\includegraphics[width=1.0\textwidth]{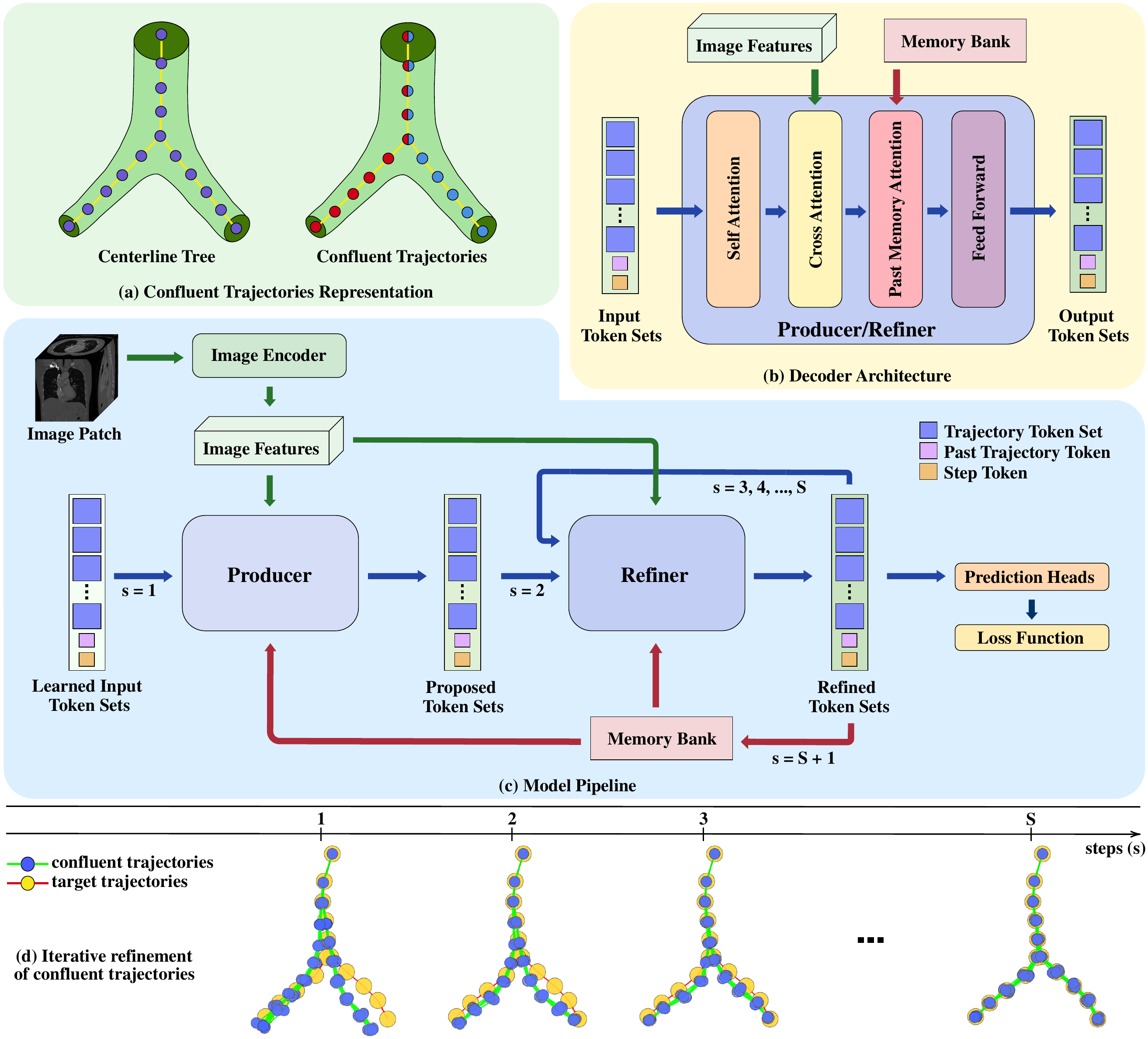}
\caption{
Overview of the RefTr architecture. \textbf{(a)} A vessel centerline tree graph with corresponding confluent trajectories and divergence position at the fifth node. \textbf{(b)} Producer/Refiner architecture with Self-, Cross-, and Past-Memory attention. \textbf{(c)} Model pipeline including Image Encoder, Producer, Refiner, and prediction heads. The Producer uses learned token sets and image features to propose multiple trajectory embeddings, which the Refiner iteratively refines over $S$ steps. Loss is computed at every step, including the Producer step (omitted for clarity). One refined token set per continuing branch is stored in the Memory Bank for the next patch. \textbf{(d)} Recurrent refinement of multiple confluent trajectories toward two target trajectories over $S$ steps.
}
\label{fig:arch}
\end{figure}

RefTr generates vascular centerline trees from CT volumes using a Transformer-based Producer–Refiner architecture and a confluent trajectory representation. The model architecture is shown in Figure~\ref{fig:arch}.

\subsection{Problem Formulation and Inference}
The goal is to infer a centerline tree graph $G=(V,E)$ from a 3D image patch. Each node $\mathbf{v}\in V$ represents a spatial position and vessel radius. RefTr represents the tree as a set of confluent trajectories $\mathcal{T}=\{T_i\}_{i=1}^{n}$, as shown in Figure~\ref{fig:arch}a, where each trajectory is a sequence of nodes originating from the patch center.

Confluence is modeled by overlapping trajectory segments until divergence positions, after which trajectories separate to form branches. During inference, predicted trajectories are converted into a tree by introducing bifurcations at divergence points and terminations at predicted end positions. Branches extending beyond trajectory length are explored recursively using new patches.

\subsection{RefTr architecture}
RefTr consists of an image encoder and a customized Transformer decoder with two modules, a Producer and a Refiner. The encoder extracts feature representations from the input patch. For a fair comparison, we adopt the same encoder as Trexplorer Super~\cite{naeem2025trexplorer}, a modified SwinUNETR~\cite{hatamizadeh2021swin}. The Producer generates $n$ initial trajectory proposals from learned token sets, and the Refiner recurrently updates them over $S-1$ steps using shared parameters. Both modules use a single decoder block. This recurrent process improves alignment with the $m$ target trajectories ($n > m$) while keeping the model parameter efficient. The full pipeline is shown in Figure~\ref{fig:arch}c.

Each trajectory token set includes $L$ node tokens that predict node positions and radii, an end token that predicts the termination position, and a divergence token that is pairwise concatenated with those of other trajectories to predict branching positions between them. The end and divergence positions are predicted as normalized values along the trajectory and are discretized to node indices in $\{0,\dots,L-1\}$.

Self-attention models interactions among trajectories, enabling effective many-to-one matching in which predictions are distributed approximately evenly across targets, yielding an ensembling effect that improves recall and precision. Cross-attention aggregates image features. Past-Memory attention incorporates context from a memory bank that stores token sets of ancestor trajectories from up to five previously processed patches. After each patch, one refined trajectory token set per continuing branch is stored in the memory bank and used as past context for the next patch. This context helps maintain the correct trajectory direction, especially for large vessels where a single patch has limited field of view, and reduces duplicate predictions across patches. As in Trexplorer Super, each training sample contains six patches along a single trajectory, so the memory bank holds up to five ancestor token sets. The detailed decoder architecture is shown in Figure~\ref{fig:arch}b.

In addition to trajectory token sets, we use a past trajectory token that encodes up to the last ten positions and radii from the patch center. This token helps determine the centerline direction, especially when the memory bank is empty. We also include a step token that indicates the current refinement step and regulates the magnitude of updates made by the Refiner.

\subsection{Matching and Loss}
Predicted trajectories are matched to targets using a combined L1 (Manhattan) cost over position and radius differences. To allow multiple predictions per branch, we employ many-to-one matching by replicating targets and solving a bipartite assignment with the Hungarian algorithm~\cite{kuhn1955hungarian}.

All loss components use L1 loss. The total loss aggregates trajectory position and radius errors across refinement steps, and includes end-position and divergence losses at the final step, where trajectories are most refined:
\begin{equation}
\mathcal{L} = \sum_{s} (\alpha_{\text{pos}}\mathcal{L}_{\text{pos}}^{(s)} + \alpha_{\text{rad}}\mathcal{L}_{\text{rad}}^{(s)})
+ \alpha_{\text{end}}\mathcal{L}_{\text{end}} + \alpha_{\text{div}}\mathcal{L}_{\text{div}}.
\end{equation}

To ensure consistent supervision, the matching is computed only once using the predictions from the first step. This fixed assignment is reused for all remaining $S - 1$ refinement steps during loss computation.

\subsection{Tree Non-Maximum Suppression}
RefTr predicts multiple trajectories per target using many-to-one matching, creating an ensembling effect that maximizes recall and improves precision. Duplicate predictions are filtered at the patch level using the pairwise divergence position. To remove duplicates globally, we introduce Tree Non-Maximum Suppression (TNMS), an efficient algorithm that merges overlapping branches while preserving tree topology. 

Given a predicted tree with $N$ nodes, TNMS performs a pre-order traversal and identifies duplicate nodes using a radius-adaptive spatial threshold. Branches with a high fraction of duplicates are merged, and any cycles are resolved by retaining edges closest to the root. Using KD-tree queries, the algorithm runs in $\mathcal{O}(N \log N)$ time, processing graphs with thousands of nodes in under one second on a CPU.

\section{Experiments and Results}

\subsection{Datasets}
We evaluate on the benchmark centerline datasets~\cite{naeem2025trexplorersuper} introduced in prior work~\cite{naeem2025trexplorer}, including a synthetic dataset, the ATM'22 airway dataset~\cite{zhang2023multi}, and the Parse 2022 pulmonary artery dataset~\cite{luo2023efficient}, covering diverse structures.

\subsection{Evaluation metrics}
Following Trexplorer Super~\cite{naeem2025trexplorer}, we report node-level, branch-level, and graph-level metrics. To enable a more robust evaluation, we introduce radius-aware matching thresholds that scale with the ground-truth vessel radius. A predicted node is considered a true positive if it lies within $\max(1.5, \tau^{\text{rad}} \cdot r)$ of a previously unmatched target node. Duplicate predictions are treated as false positives.

At the node-level, we report radius-aware Average Precision (rAP), Average Recall (rAR), and Average F1-score (rF1), averaged over $\tau^{\text{rad}} = 0.25{:}0.05{:}0.75$ to reduce sensitivity to threshold selection. For branch-level evaluation, a predicted branch is counted as a true positive if it overlaps with at least a fraction $\tau^{\text{match}}$ of nodes from an unmatched target branch. We report rBAP, rBAR, and rBF1, the branch-level counterparts of the node-level metrics. These are averaged over $\tau^{\text{match}} = 0.4{:}0.05{:}0.8$, with $\tau^{\text{rad}}$ fixed at $0.5$. For graph-level evaluation, we report topological metrics, specifically the mean absolute error (MAE) of Betti-0 (connected components) and Betti-1 (cycles), following Trexplorer Super.

\subsection{Implementation details}
Experiments were conducted on a single node with four NVIDIA A100 GPUs using mixed precision. RefTr was trained for 1.9 million iterations with $S=8$ refinement steps, trajectory length $L=10$, and $n=20$ predicted branches. To ensure fair comparison with Trexplorer and Trexplorer Super~\cite{naeem2025trexplorer,naeem2024trexplorer}, we used the same dataset and train–test split, adopting the authors’ reported hyperparameters (26 tokens per bifurcation node, maximum 196 tokens) and trained for 2 million iterations. For Vesselformer~\cite{prabhakar2024vesselformer}, we report the best performance obtained using the published hyperparameters, with 80 object tokens and 12 million training iterations.

\begin{table}[!t]
\centering
\caption{Comparison of various models using point-level metrics.}
\label{tab:res-table-1}
{\fontsize{8}{8.5}\selectfont
\begin{tabular}{c|c|c|c|c}
\toprule
\multirow{2}{*}{Model} & \multicolumn{4}{c}{Point-level${@\,\tau^{\text{rad}}=[0.25{:}0.05{:}0.75]}$} \\ 
                       & rAP(\%)$\uparrow$ & rAR(\%)$\uparrow$ & rF1(\%)$\uparrow$ & Radius (MAE)$\downarrow$ \\ 
\midrule
\multicolumn{5}{c}{Synthetic Dataset} \\ 
\midrule
Vesselformer        & 48.53 ± 10.35  & 70.05 ± 2.76  & 53.15 ± 8.19 & 0.43 ± 0.011 \\
Trexplorer          & 31.10 ± 9.52  & 78.64 ± 4.10  & 39.64 ± 8.68 & 0.23 ± 0.032 \\
Trexplorer Super & 92.10 ± 3.29  & 70.58 ± 3.03  & 77.98 ± 1.90 & \textbf{0.10 ± 0.006} \\ 
RefTr (ours)   &  \textbf{94.88 ± 0.77}  & \textbf{79.39 ± 1.03}  & \textbf{85.69 ± 1.30} & 0.45 ± 0.093 \\
\cmidrule(lr){1-5}
\multicolumn{5}{c}{ATM'22 Dataset} \\ 
\midrule
Vesselformer        & 24.85 ± 1.47  & 38.48 ± 1.00  & 29.84 ± 1.37 & 0.79 ± 0.020 \\
Trexplorer          & 4.54 ± 1.44   & 5.90 ± 0.70   & 4.49 ± 0.29  & 0.98 ± 0.086 \\
Trexplorer Super    & 71.32 ± 1.22  & 60.84 ± 3.38  & 61.45 ± 1.23 & \textbf{0.40 ± 0.021} \\ 
RefTr (ours)   &  \textbf{74.92 ± 1.17}  & \textbf{66.73 ± 1.48}  & \textbf{68.15 ± 0.59} & 0.60 ± 0.036 \\
\cmidrule(lr){1-5}
\multicolumn{5}{c}{Parse 2022 Dataset} \\ 
\midrule
Vesselformer        & 21.93 ± 2.06  & 18.21 ± 1.00  & 19.54 ± 2.69 & 1.11 ± 0.027 \\
Trexplorer          & 11.37 ± 3.73   & 13.42 ± 7.80  & 11.36 ± 5.17 & 1.22 ± 0.304 \\
Trexplorer Super    & 53.65 ± 5.91  & 35.91 ± 3.22  & 40.01 ± 3.86 & \textbf{0.58 ± 0.025} \\
RefTr (ours)   &  \textbf{53.68 ± 1.09}  & \textbf{36.69 ± 1.12}  & \textbf{42.37 ± 1.13} & 0.75 ± 0.054 \\
\bottomrule
\end{tabular}}
\end{table}

\begin{table}[!t]
\centering
\caption{Comparison of various models using branch-level and graph-level metrics.}
\label{tab:res-table-2}
{\fontsize{8}{8.5}\selectfont
\begin{tabular}{c|c|c|c|c|c}
\toprule
\multirow{2}{*}{Model} &  \multicolumn{3}{c|}{Branch-level${@\,\tau^{\text{match}}=[0.4{:}0.05{:}0.8]}$} & \multicolumn{2}{c}{Graph-level (MAE)} \\ 
                        & rBAP(\%)$\uparrow$ & rBAR(\%)$\uparrow$ & rBF1(\%)$\uparrow$  & Betti-0$\downarrow$ & Betti-1$\downarrow$ \\ 
\midrule
\multicolumn{6}{c}{Synthetic Dataset} \\ 
\midrule
Vesselformer               & 20.43 ± 4.22 & 43.63 ± 4.72  & 25.43 ± 3.75 & 81.7 ± 16.8   & 653.5 ± 138.7  \\
Trexplorer                 & 30.58 ± 8.66 & 77.47 ± 4.11 & 39.40 ± 7.82 & \textbf{0.00 ± 0.0}   & \textbf{0.00 ± 0.0}  \\
Trexplorer Super           & 92.35 ± 3.54 & 70.82 ± 3.06      & 78.22 ± 1.78  & \textbf{0.00 ± 0.0}   & \textbf{0.00 ± 0.0}  \\ 
RefTr (ours)   &  \textbf{96.75 ± 0.72}  & \textbf{79.96 ± 1.15}  & \textbf{86.76 ± 1.38} & \textbf{0.00 ± 0.0} & \textbf{0.00 ± 0.0} \\
\cmidrule(lr){1-6}
\multicolumn{6}{c}{ATM'22 Dataset} \\ 
\midrule
Vesselformer               & 11.70 ± 0.73 & 16.77 ± 1.07  & 13.46 ± 0.81 & 312.5 ± 25.1   & 180.4 ± 35.9  \\
Trexplorer                 & 4.02 ± 1.71 & 2.06 ± 0.46 & 2.33 ± 0.31 & \textbf{0.00 ± 0.0}   & \textbf{0.00 ± 0.0}  \\
Trexplorer Super           & 68.25 ± 1.75 & 60.91 ± 3.64  & 59.87 ± 1.39  & \textbf{0.00 ± 0.0}   & \textbf{0.00 ± 0.0}  \\ 
RefTr (ours)   &  \textbf{68.40 ± 1.97}  & \textbf{66.41 ± 1.53}  & \textbf{64.23 ± 0.89} & \textbf{0.00 ± 0.0} & \textbf{0.00 ± 0.0} \\
\cmidrule(lr){1-6}
\multicolumn{6}{c}{Parse 2022 Dataset} \\ 
\midrule
Vesselformer               & 9.23 ± 0.80 & 6.31 ± 0.85  & 7.30 ± 0.45 & 410.1 ± 23.9   & 246.7 ± 78.1 \\
Trexplorer                 & 11.32 ± 3.56 & 11.77 ± 7.22 & 10.45 ± 4.98 & \textbf{0.00 ± 0.0}   & \textbf{0.00 ± 0.0}  \\
Trexplorer Super           & \textbf{51.42 ± 5.58} & \textbf{34.41 ± 1.09} & \textbf{38.17 ± 3.63}  & \textbf{0.00 ± 0.0}   & \textbf{0.00 ± 0.0}  \\
RefTr (ours)   &  47.17 ± 1.19  & 32.31 ± 1.13  & 37.09 ± 1.16 & \textbf{0.00 ± 0.0} & \textbf{0.00 ± 0.0} \\
\bottomrule
\end{tabular}}
\end{table}

\subsection{Results}

Table~\ref{tab:res-table-1} and Table~\ref{tab:res-table-2} report the mean and standard deviation over five runs for point-level, branch-level, and graph-level metrics. At the point level, RefTr achieves the highest recall and precision across all datasets with sub-voxel radius error. On Parse 2022, improvements are smaller due to missing small branches in the ground truth, which RefTr correctly predicts (as verified by a radiologist review) but is penalized for, and encoder limitations from limited training data (72 training examples). We used the same encoder as Trexplorer Super for a fair comparison.

RefTr achieves superior branch-level metrics on the Synthetic and ATM'22 datasets. On Parse 2022, performance is comparable. However, the small branches that are missing in the ground truth have a larger negative impact at this level because small and large branches are weighted equally. RefTr also predicts topologically correct centerline trees, as reflected by the graph-level metrics. 

\begin{figure*}[!t]
\centering
\includegraphics[width=0.99\textwidth]{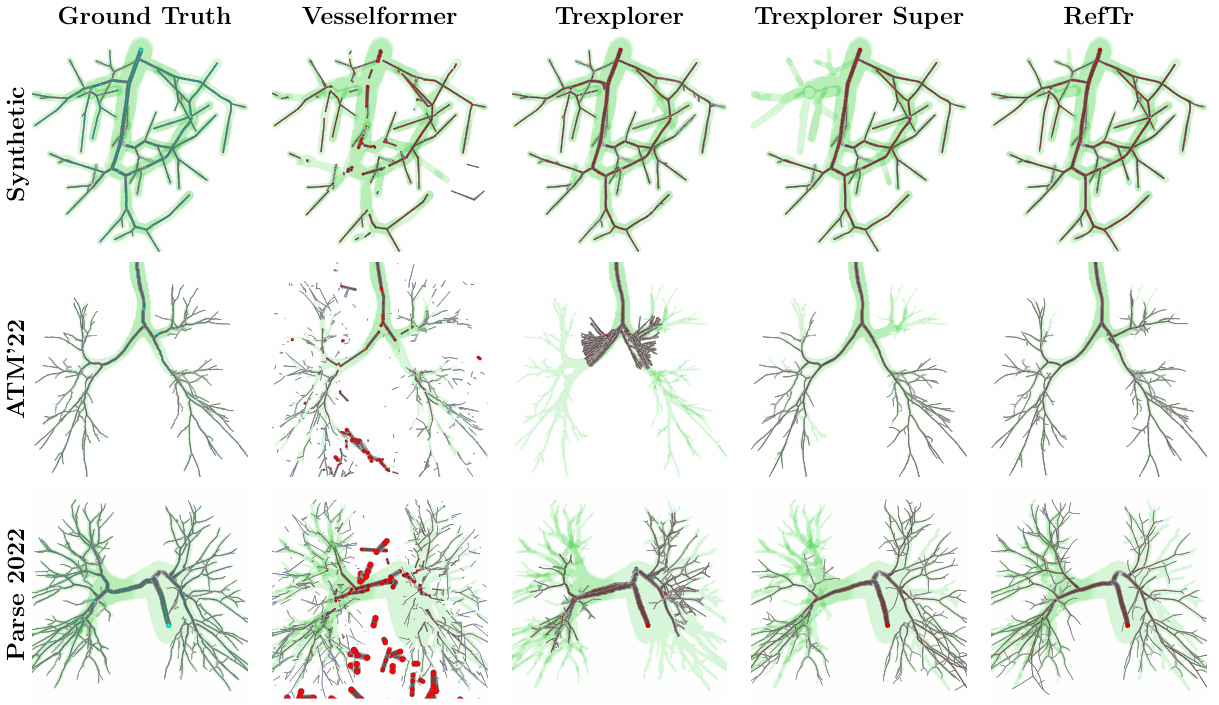}
\caption{Visual comparison of ground truth and predictions from RefTr and baseline methods on one sample per dataset. Marker sizes are proportional to node radii. The vessel mask (green) is for visualization only and not used during training or evaluation.}
\label{fig:comp}
\end{figure*}

Table~\ref{tab:param-table} shows that, compared to the Trexplorer Super, RefTr reduces the number of decoder parameters by 2.4$\times$, achieves faster inference, and requires substantially less training time and memory, while delivering superior performance. Figure~\ref{fig:comp} provides a qualitative comparison.

\begin{table}[]
\setlength{\tabcolsep}{0.45mm}
\centering
\caption{Number of parameters, runtime, training time, and memory usage for the compared models. Note: Vesselformer has a much higher effective runtime due to a multi-hour post-processing. Experiments conducted on four Nvidia A100 80GB GPUs.}
\label{tab:param-table}
{\fontsize{8}{8.5}\selectfont
\begin{tabular}{c|c|c|c|c|c|c}
\toprule
\multirow{2}{*}{Model} & \multicolumn{3}{c|}{Parameters (M)} & \multirow{2}{*}{\shortstack{Runtime \\ (ms/patch)}} & \multirow{2}{*}{\shortstack{Training\\ (hrs)}} & \multirow{2}{*}{\shortstack{GPU Mem.\\ (GBs)}}\\ 
                       & Encoder & Decoder & Total & & & \\ 
\midrule
Vesselformer & 82.83 & 13.59 & 98.73 & $23^*$ & 37.4 & \textbf{13.8}\\
Trexplorer & 23.31 & 25.22 & 55.46 & 222 & 18.6 & 51.34 \\
Trexplorer Super & 23.12 & 25.22 & 55.27 & 189 & 37.2 & 73.9\\
RefTr (ours) & \textbf{23.12} & \textbf{10.50} & \textbf{42.32} & \textbf{156} & \textbf{18.2} & 39.5 \\
\bottomrule
\end{tabular}}
\end{table}

\begin{table}[!t]
\centering
\caption{Ablation study evaluating the impact of the key components and design choices for the RefTr model on the ATM'22 dataset.}
\label{tab:ablation-table}
{\fontsize{8}{9}\selectfont
\begin{tabular}{c|c|c|c}
\toprule
\multirow{2}{*}{Description} & \multicolumn{3}{c}{Point-level${@\,\tau^{\text{rad}}=[0.25{:}0.05{:}0.75]}$} \\ 
&  rAP(\%)$\uparrow$ & rAR(\%)$\uparrow$ & rF1(\%)$\uparrow$ \\ 
\midrule
\multicolumn{4}{c}{Key High-level Components} \\ 
\cmidrule(lr){1-4}
No Recurrent Refinement ($S=2$) & 61.96 ± 2.17 & 58.08 ± 1.17 & 56.70 ± 1.66 \\
No Many-to-one Matching & 66.70 ± 1.27 & 63.25 ± 3.49 & 61.34 ± 1.68 \\
No Tree Non-max Suppression & 37.98 ± 2.42 & 
\textbf{69.96 ± 1.12} & 37.98 ± 3.87 \\
\cmidrule(lr){1-4}
\multicolumn{4}{c}{Recurrent Refinement Steps} \\ 
\cmidrule(lr){1-4}
Refinement Steps: $S=4$ & 65.10 ± 6.32 & 55.72 ± 2.83 & 58.17 ± 3.42 \\
Refinement Steps: $S=6$ & 64.39 ± 5.17 & 60.93 ± 2.57 & 59.42 ± 3.19 \\
Refinement Steps: $S=10$ & 71.43 ± 3.12 & 65.16 ± 1.58 & 65.60 ± 1.91 \\
\cmidrule(lr){1-4}
\multicolumn{4}{c}{Past Trajectory and Memory Attention} \\ 
\cmidrule(lr){1-4}
No Past Trajectory Token & \textbf{78.85 ± 3.24} & 37.07 ± 1.21 & 48.34 ± 2.78 \\
No Past-Memory Attention & 61.27 ± 2.12 & 66.38 ± 1.87 & 60.46 ± 2.24 \\
\cmidrule(lr){1-4}
\multicolumn{4}{c}{Producer and Refiner Design Configurations} \\ 
\cmidrule(lr){1-4}
Shared Producer and Refiner & 65.18 ± 1.44 & 61.76 ± 1.32 & 60.27 ± 1.78 \\
Increase Producer Blocks: 2 & 68.80 ± 2.24 & 62.88 ± 1.11 & 63.19 ± 1.83 \\
Increase Refiner Blocks: 2 & 71.81 ± 2.07 & 64.67 ± 1.57 & 64.46 ± 1.98 \\
\cmidrule(lr){1-4}
Proposed Model & 74.16 ± 1.45  & 67.72 ± 1.49  & \textbf{68.35 ± 1.27} \\
\bottomrule
\end{tabular}}
\end{table}

\subsection{Ablation Study}
Table~\ref{tab:ablation-table} presents an extensive ablation study of our method’s key components, including Recurrent Refinement, Many-to-One matching, Tree Non-Maximum Suppression, Past-Attention, and various Producer and Refiner design choices. We report the mean and standard deviation over three runs.

The study shows that Recurrent Refinement, Many-to-One matching, and Past-Memory attention are essential for achieving high recall, while Tree Non-Maximum Suppression is crucial for removing duplicates from the intentional overprediction. 

Using the same decoder block for both the Producer and Refiner is suboptimal because the shared block must process two different input types: learned token sets and already refined token sets. Increasing the number of Producer blocks makes the initial proposals overly rigid, which limits the Refiner’s ability to correct early errors. Conversely, increasing the number of Refiner blocks causes each refinement step to make larger updates, which can push predictions off track and make subsequent steps less able to recover. Overall, the best performance is achieved with separate Producer and Refiner modules, each using a single decoder block and $S=8$ refinement steps.

\section{Conclusion}
We present RefTr, a parameter-efficient framework for generating topologically accurate vascular centerline trees from 3D medical images. By predicting confluent trajectories and refining them recurrently, RefTr delivers state-of-the-art performance with substantially improved efficiency. Results across multiple datasets show faster inference and strong potential for clinical vascular analysis. Future work will focus on improving the image encoder to further enhance feature quality and performance on challenging cases.

\bigskip

\noindent \textbf{Acknowledgments.} Compute and storage resources were provided by NAISS (grant 2022-06725, Swedish Research Council) and the Berzelius system at the National Supercomputer Centre, funded by the Knut and Alice Wallenberg Foundation.

\bibliographystyle{splncs04}
\bibliography{references}

\clearpage
\appendix
\renewcommand{\thesection}{\Alph{section}}
\section*{Supplementary Material}

This supplementary material is organized as follows.
Section~A describes how we reconstruct a centerline tree from the predicted confluent trajectories.
Section~B summarizes the experimental setup, including the architecture and training hyperparameters.
Section~C details the matching procedure and training losses.
Section~D describes Tree Non-Max Suppression (TNMS).
Section~E presents additional qualitative examples and discusses annotation ambiguities and representative failure cases.

\section{From Confluent Trajectories to a Centerline Tree}

RefTr predicts a set of confluent trajectories, together with an end position for each trajectory and a divergence position for each pair of trajectories.
These positions are represented as continuous values in $[0,1]$.
We first unnormalize them to the trajectory index range $[0, L-1]$ and then discretize them by rounding to the nearest integer, yielding a trajectory node index in $\{0, 1, \dots, L-1\}$ at which the end or divergence occurs.

We construct the centerline tree from the predicted confluent trajectories as follows:

\begin{enumerate}
\item Group trajectories by their predicted divergence positions, such that trajectory pairs with divergence position $L-1$ are assigned to the same group.
A divergence position of $L-1$ indicates that the two trajectories remain confluent until the final node.
\item Merge the trajectories within each group into a representative trajectory by averaging their node positions and node radii, and by averaging their end positions.
\item For the representative trajectories, compute pairwise divergence positions by averaging the corresponding predicted divergence positions.
\item Starting from the shared root, construct the tree level by level in a breadth-first manner.
\item At each level, cluster the representative trajectories according to their divergence positions.
When multiple clusters are present, create a new branch for each cluster.
These clusters may split again at later levels, giving rise to further branches.
\item Terminate branches once their predicted end positions are reached.
\end{enumerate}

This procedure produces a single connected, acyclic tree and therefore preserves topological correctness.

\section{Experimental Setup}

We provide the key architecture and training hyperparameters in Tables~\ref{tab:arch_hyperparams} and~\ref{tab:train_hyperparams}, respectively.

\begin{table}[]
\centering

\begin{minipage}[t]{0.48\columnwidth}
\centering
\setlength{\tabcolsep}{2.2mm}
\caption{Architecture hyperparameters.}
\label{tab:arch_hyperparams}
{\fontsize{9}{11}\selectfont
\begin{tabular}{@{}ll@{}}
\toprule
\textbf{Hyperparameter} & \textbf{Value} \\
\midrule
3D patch size  & $64\times64\times64$ \\
Producer blocks & 1 \\
Refiner blocks & 1 \\
Decoder heads & 16 \\
Token dimension & 512 \\
Decoder FF dim & 2048 \\
Position head dim & [512,512,512] \\
Radius head dim & [512,512,512] \\
Divergence head dim & [1024,512,256,128] \\
End head dim & [512,256,128,64] \\
Encoder feat. dim & 24 \\
Encoder depths & [2,2,2,2] \\
Encoder heads & [3,6,12,24] \\
Encoder patch size & 2 \\
Encoder window size & 7 \\
\bottomrule
\end{tabular}
}
\end{minipage}
\hspace{0.01\columnwidth}
\begin{minipage}[t]{0.48\columnwidth}
\centering
\setlength{\tabcolsep}{2.2mm}
\caption{Training hyperparameters.}
\label{tab:train_hyperparams}
{\fontsize{9}{11}\selectfont
\begin{tabular}{@{}ll@{}}
\toprule
\textbf{Hyperparameter} & \textbf{Value} \\
\midrule
Training iterations & 1.92M \\
Sequence length $L$ & 10 \\
Refinement steps $S$ & 8 \\
Predicted branches $n$ & 20 \\
Position loss weight $\alpha_{\text{pos}}$ & 4.2 \\
Position cost weight $\lambda_{\text{pos}}$ & 3 \\
Radius loss weight $\alpha_{\text{rad}}$ & 1.15 \\
Radius cost weight $\lambda_{\text{rad}}$ & 1 \\
End loss weight $\alpha_{\text{end}}$ & 0.94 \\
Divergence loss weight $\alpha_{\text{div}}$ & 0.3 \\
\bottomrule
\end{tabular}
}
\end{minipage}

\end{table}

\section{Matching and Loss Function}

\subsection{Matching cost}

Given $n$ predicted trajectories $\hat{\mathcal{T}} = \{\hat{T}_i\}_{i=1}^n$ and $m$ target trajectories $\mathcal{T} = \{T_j\}_{j=1}^m$, we compute a matching from predictions to targets using a cost matrix $\mathbf{C} \in \mathbb{R}^{n \times m}$.
Each entry $\mathbf{C}_{i,j}$ is defined as a weighted sum of position and radius costs:
\begin{equation}
\mathbf{C}_{i,j}
=
\lambda_{\text{pos}} \cdot \mathcal{C}_{\text{pos}}(\hat{T}_i, T_j)
+
\lambda_{\text{rad}} \cdot \mathcal{C}_{\text{rad}}(\hat{T}_i, T_j),
\end{equation}
where
\begin{align}
\mathcal{C}_{\text{pos}}(\hat{T}_i, T_j)
&=
\frac{1}{L}
\sum_{l=1}^L
\left\|
\hat{\mathbf{x}}_i^{(l)} - \mathbf{x}_j^{(l)}
\right\|_1, \\
\mathcal{C}_{\text{rad}}(\hat{T}_i, T_j)
&=
\frac{1}{L}
\sum_{l=1}^L
\left|
\hat{r}_i^{(l)} - r_j^{(l)}
\right|.
\end{align}
Here, $\lambda_{\cdot}$ denotes the scalar weight of each cost component.

To ensure consistent supervision across $S$ refinement steps, the matching is computed only once using the predictions from the first decoding step.
This fixed assignment is then reused for the remaining $S-1$ steps during loss computation.

\subsection{Many-to-one matching}

We choose $n > m$ and enable many-to-one matching by replicating the target set to form a square cost matrix $\tilde{\mathbf{C}} \in \mathbb{R}^{n \times n}$.
This allows us to apply the Hungarian algorithm to perform bipartite matching between the $n$ predicted trajectories and the $n$ replicated target trajectories:
\begin{equation}
\sigma^* = \arg\min_{\sigma \in \mathfrak{S}_n} \sum_{i=1}^n \tilde{\mathbf{C}}_{\sigma(i), i},
\end{equation}
where $\mathfrak{S}_n$ denotes the set of all permutations over $n$ elements and $\sigma^*$ is the permutation that minimizes the total cost.

Since the target trajectories are replicated, we map the matched replicated index back to the original target index set $\{1,\dots,m\}$:
\begin{equation}
\hat{a}_i = 1 + ((\sigma^*(i)-1) \bmod m),
\end{equation}
where $\hat{a}_i$ denotes the target trajectory assigned to the $i$-th predicted trajectory.

\subsection{Loss components}

Let $\hat{a}_i$ denote the target trajectory assigned to predicted trajectory $i$.
For each decoder step $s = 1, \dots, S$, we define the position and radius losses as
\begin{align}
\mathcal{L}_{\text{pos}}^{(s)}
&=
\frac{1}{nL}
\sum_{i=1}^n
\sum_{l=1}^L
\left\|
\hat{\mathbf{x}}_i^{(l,s)} - \mathbf{x}_{\hat{a}_i}^{(l)}
\right\|_1, \\
\mathcal{L}_{\text{rad}}^{(s)}
&=
\frac{1}{nL}
\sum_{i=1}^n
\sum_{l=1}^L
\left|
\hat{r}_i^{(l,s)} - r_{\hat{a}_i}^{(l)}
\right|.
\end{align}

The end loss and divergence loss are computed only at the final decoding step $s = S$, where the predicted trajectories are most refined.
The end loss is
\begin{align}
\mathcal{L}_{\text{end}}
=
\frac{1}{n}
\sum_{i=1}^n
\left|
\hat{e}_i^{(S)} - e_{\hat{a}_i}
\right|.
\end{align}

Let $\mathbf{D} \in \mathbb{R}^{m \times m}$ denote the target divergence matrix, where $\mathbf{D}_{j,\ell}$ is the divergence position between target trajectories $T_j$ and $T_\ell$.
Given the assignment $\hat{a}_i$ for each predicted trajectory $i$, we construct a prediction-aligned target divergence matrix $\mathbf{D}' \in \mathbb{R}^{n \times n}$ by
\begin{align}
\mathbf{D}'_{i,k} = \mathbf{D}_{\hat{a}_i,\hat{a}_k}, \quad \forall i \neq k.
\end{align}
That is, the target divergence supervising the pair of predictions $(i,k)$ is taken from the divergence between their assigned target trajectories.

The divergence loss is then defined as
\begin{align}
\mathcal{L}_{\text{div}}
=
\frac{1}{n(n-1)}
\sum_{\substack{i,k=1 \\ i \neq k}}^n
\left|
\hat{\mathbf{D}}_{i,k}^{(S)} - \mathbf{D}'_{i,k}
\right|.
\end{align}

\subsection{Total loss}

The overall loss aggregates the position and radius losses across all steps and adds the end and divergence losses from the final step:
\begin{equation}
\begin{split}
\mathcal{L}
=
\sum_{s=1}^S
\left(
\alpha_{\text{pos}} \mathcal{L}_{\text{pos}}^{(s)}
+
\alpha_{\text{rad}} \mathcal{L}_{\text{rad}}^{(s)}
\right)
+
\alpha_{\text{end}} \mathcal{L}_{\text{end}}
+
\alpha_{\text{div}} \mathcal{L}_{\text{div}},
\end{split}
\label{eq:total-loss}
\end{equation}
where $\alpha_{\cdot}$ are scalar weights that balance the contributions of the different loss terms.

\section{Tree Non-Max Suppression}

Tree Non-Max Suppression (TNMS) is an efficient post-processing algorithm that merges overlapping branches in predicted trees while preserving topological consistency.
Although the divergence head resolves many duplicates, TNMS removes any remaining overlaps.

Given a directed tree $G = (V, E)$ with 3D node positions and radii, TNMS performs a top-down (pre-order) traversal from the root.
It maintains a list of visited nodes and uses a KD-tree for fast spatial queries.
The spatial matching threshold is adaptive: for a node $v$ with radius $r_v$, we use
\[
\tau = \max(\tau_{\text{min}}, \tau_{\text{pos}} \cdot r_v),
\]
with $\tau_{\text{pos}} = 0.3$ and $\tau_{\text{min}} = 2.0$.
The radius-aware threshold allows TNMS to operate robustly across vascular trees of different scales.
Nodes within this threshold of previously visited nodes are flagged as duplicates.

Branches with a sufficiently large fraction of flagged nodes (above $\rho = 0.2$) are marked for merging.
The hyperparameters $\tau_{\text{pos}}$, $\tau_{\text{min}}$, and $\rho$ were selected by simple grid search.
Duplicate nodes are grouped and replaced by a single node, and cycles are removed by retaining only the incoming edge whose parent is closest to the root.
The full procedure is summarized in Algorithm~\ref{alg:tree-nms}.

TNMS is fast and scalable.
It requires a single traversal, uses sublinear KD-tree lookups, and compares each node only to previously visited nodes, yielding $\mathcal{O}(N \log N)$ complexity for $N$ nodes.
Merging and cycle resolution are lightweight, allowing TNMS to process graphs with thousands of nodes in under a second on CPU.

\begin{algorithm}[t]
\caption{Tree Non-Max Suppression}
\label{alg:tree-nms}
\begin{algorithmic}[1]
\REQUIRE Graph $G = (V, E)$, radius scaling threshold $\tau_{\text{pos}}$, minimum distance threshold $\tau_{\text{min}}$, duplicate ratio $\rho$
\ENSURE Deduplicated graph $G$
\STATE Initialize visited nodes $\mathcal{V} \gets []$, positions $\mathcal{P} \gets []$
\STATE Initialize branch nodes $\mathcal{B}_v \gets []$, positions $\mathcal{B}_p \gets []$
\STATE Set root node $v_{\text{root}}$
\FOR{each node $v$ in pre-order from $v_{\text{root}}$}
    \IF{$v$ starts a new branch}
        \STATE Append $\mathcal{B}_v, \mathcal{B}_p$ to $\mathcal{V}, \mathcal{P}$; rebuild KD-tree
        \STATE Reset $\mathcal{B}_v \gets []$, $\mathcal{B}_p \gets []$
    \ENDIF
    \STATE Append $v$ and its position to $\mathcal{B}_v, \mathcal{B}_p$
    \STATE Query KD-tree for nearest neighbor $u$ within $\tau = \max(\tau_{\text{min}}, \tau_{\text{pos}} \cdot r_v)$
    \IF{match found}
        \STATE Mark $v$ as duplicate of $u$
    \ENDIF
\ENDFOR
\STATE Identify branches with duplicate ratio $\geq \rho$
\STATE Group duplicate pairs into sets and merge each set into a single node
\FOR{each node $v$ with in-degree $> 1$}
    \STATE Keep only the edge from the parent closest to the root
\ENDFOR
\RETURN $G$
\end{algorithmic}
\end{algorithm}

\section{Failure Cases and Annotation Ambiguities}
In this section, we present qualitative cases from the PARSE 2022 dataset that illustrate annotation ambiguities affecting evaluation, as well as representative prediction failures. The goal is to distinguish genuine failure cases from disagreements that arise from incomplete or ambiguous reference annotations.

\subsection{Missing Ground-Truth Annotations}

\begin{figure}[!ht]
\centering
\includegraphics[width=1.0\textwidth]{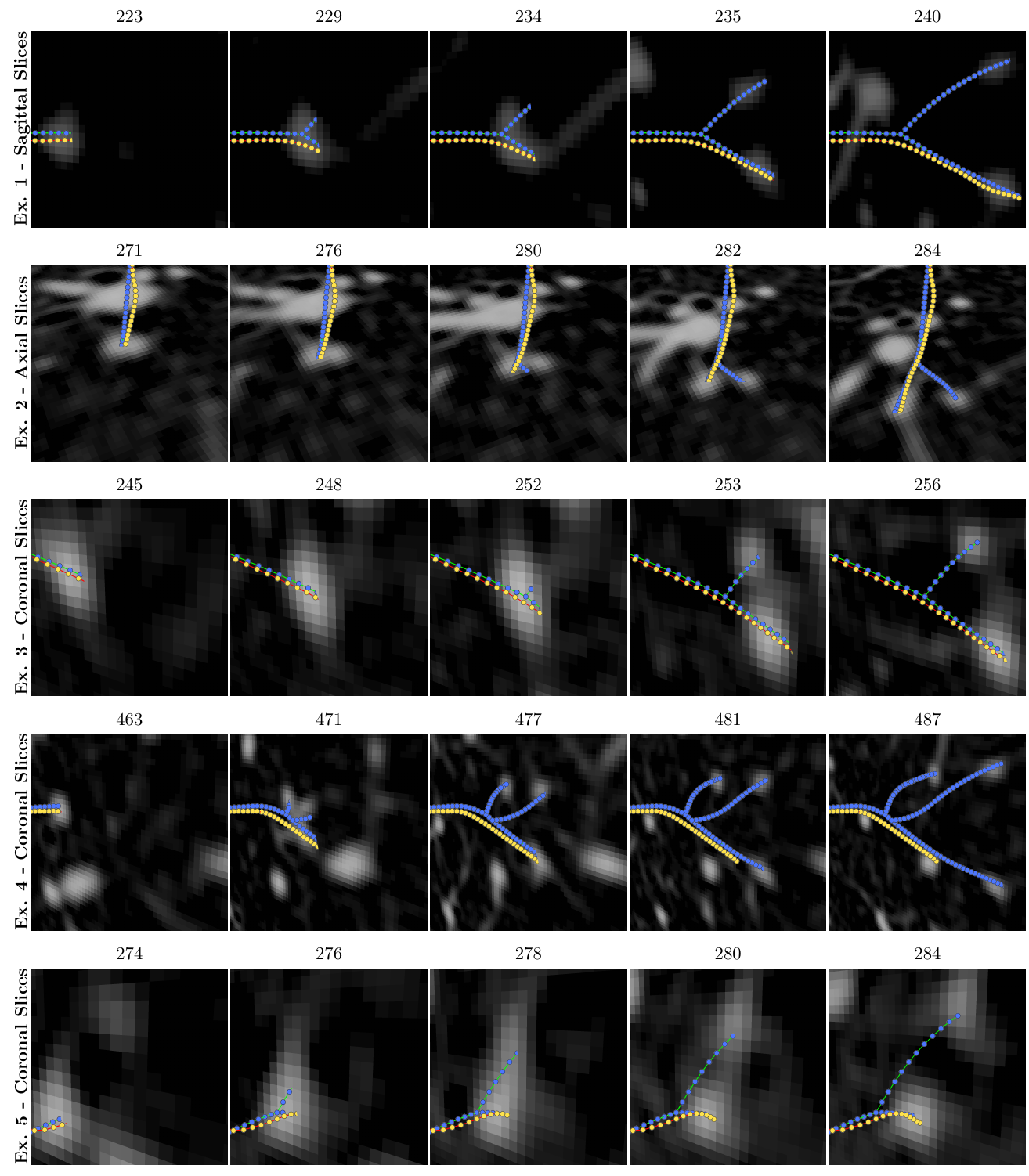}
\caption{Qualitative examples of vessel bifurcations missing in the ground truth (yellow) but correctly predicted by RefTr (blue). Each row shows one example across a sequence of CT image slices. Only vessels branching from the selected vessel in the first slice are shown.}
\label{fig:bif_miss}
\end{figure}

\begin{figure}[!ht]
\centering
\includegraphics[width=1.0\textwidth]{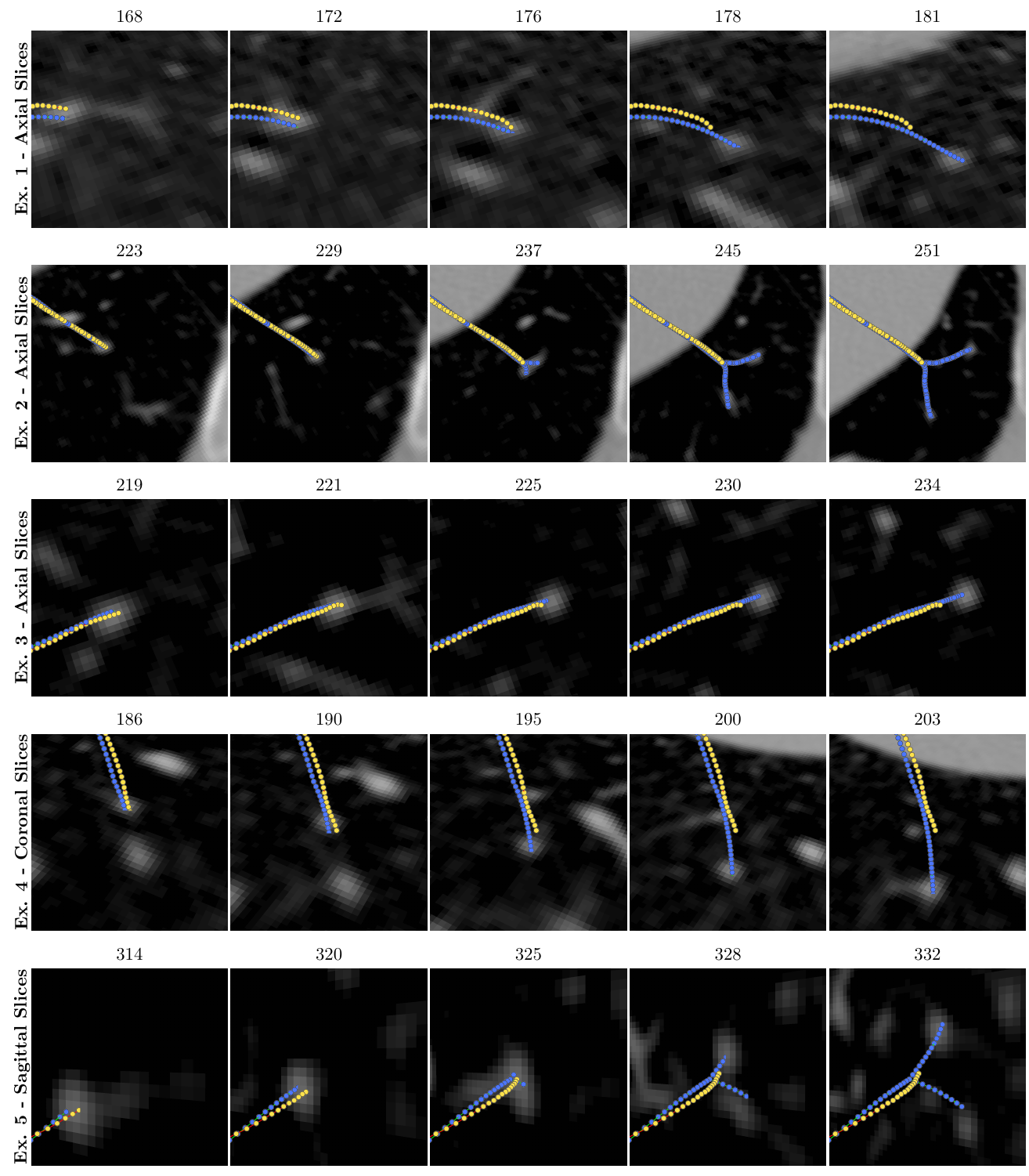}
\caption{Qualitative examples of vessels terminated early in the ground truth (yellow) but correctly predicted by RefTr (blue). Each row shows one example across a sequence of CT image slices. Only vessels branching from the selected vessel in the first slice are shown.}
\label{fig:early_term}
\end{figure}

Figure~\ref{fig:bif_miss} shows cases in which a bifurcation occurs along the selected vessel and a new branch becomes visible in the CT slice sequence, but the ground-truth annotation does not follow that branch, whereas RefTr does.
Similarly, Figure~\ref{fig:early_term} shows examples in which the ground-truth annotation terminates early even though the vessel continues to be visible in the CT data and is followed by RefTr.

In both cases, the RefTr prediction remains anatomically plausible, but the disagreement is counted as an error under the reference annotation.
As a result, these cases reduce the reported precision and F1 score in both point-level and branch-level evaluation.

These missing annotations were confirmed by an expert radiologist.
The PARSE 2022 dataset paper does not specify explicit criteria for excluding vessels from annotation.
Although many datasets omit vessels below a certain diameter, this does not appear to fully explain the examples shown here, since PARSE 2022 emphasizes that thinner pulmonary arteries are both more difficult to detect and clinically important.
Depending on the application, the predicted centerline graphs can be post-processed to filter branches using length or radius thresholds.

\subsection{Earlier Predicted Bifurcations}

\begin{figure}[!ht]
\centering
\includegraphics[width=1.0\textwidth]{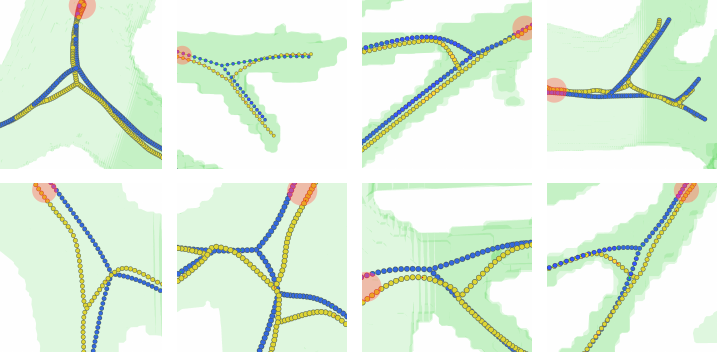}
\caption{Qualitative examples of differing bifurcation locations between the ground truth (yellow) and RefTr predictions (blue). The red marker denotes the starting point of the highlighted vessel within each patch. Each grid cell shows a separate case. RefTr bifurcates earlier to avoid missing a branch.}
\label{fig:bif_diff}
\end{figure}

Figure~\ref{fig:bif_diff} shows cases in which RefTr predicts a bifurcation slightly earlier than the ground-truth annotation.
This behavior likely reflects a preference for predicting divergence conservatively so as not to miss an emerging branch.

Both the predicted and reference centerlines remain spatially and topologically plausible, since the exact location of a centerline node or bifurcation is inherently ambiguous.
This ambiguity becomes more pronounced for larger vessels.
Under branch-level evaluation, such small shifts can still reduce the reported score, especially at stricter matching thresholds ($\tau^{\text{match}} > 0.7$).

\subsection{Failure Cases}

\begin{figure}[!ht]
\centering
\includegraphics[width=1.0\textwidth]{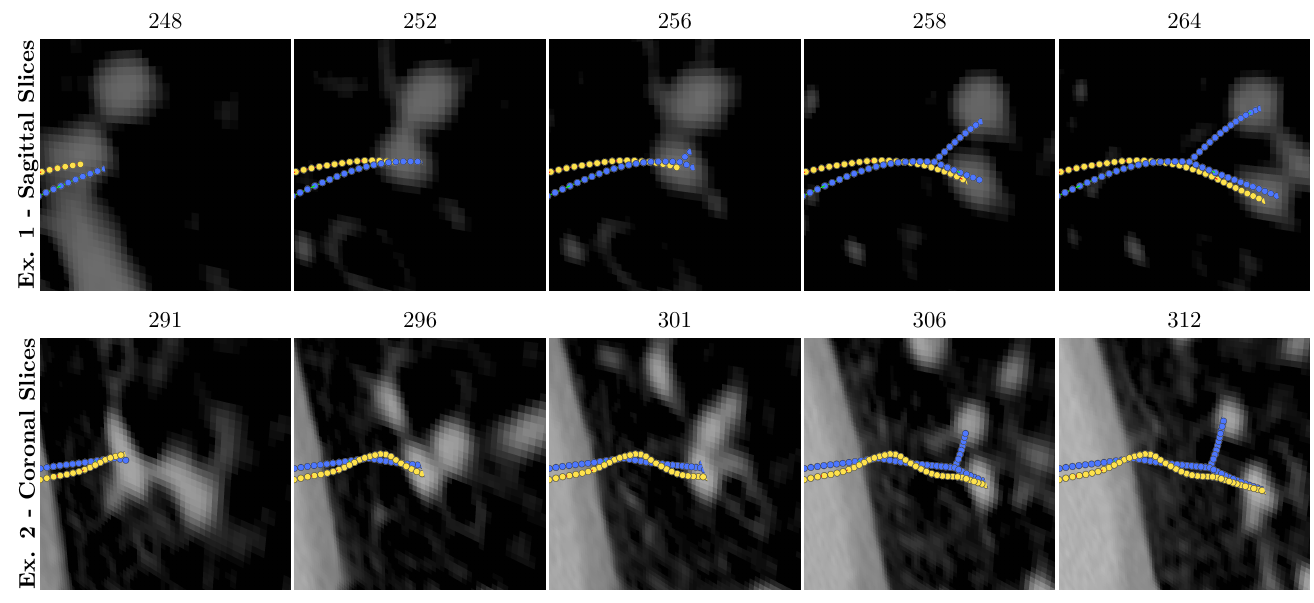}
\caption{Qualitative examples of failure cases where two nearby branches run close together before diverging. At the divergence point, RefTr incorrectly begins to track the diverging branch as a new branch.}
\label{fig:near_miss}
\end{figure}

Figure~\ref{fig:near_miss} shows cases in which another vessel comes very close to the currently tracked vessel, either touching or nearly touching before diverging.
At the divergence point, RefTr interprets the nearby vessel as a new branch and begins to track it.
Under the current annotation, this additional branch is counted as a false positive.

\begin{figure}[!ht]
\centering
\includegraphics[width=1.0\textwidth]{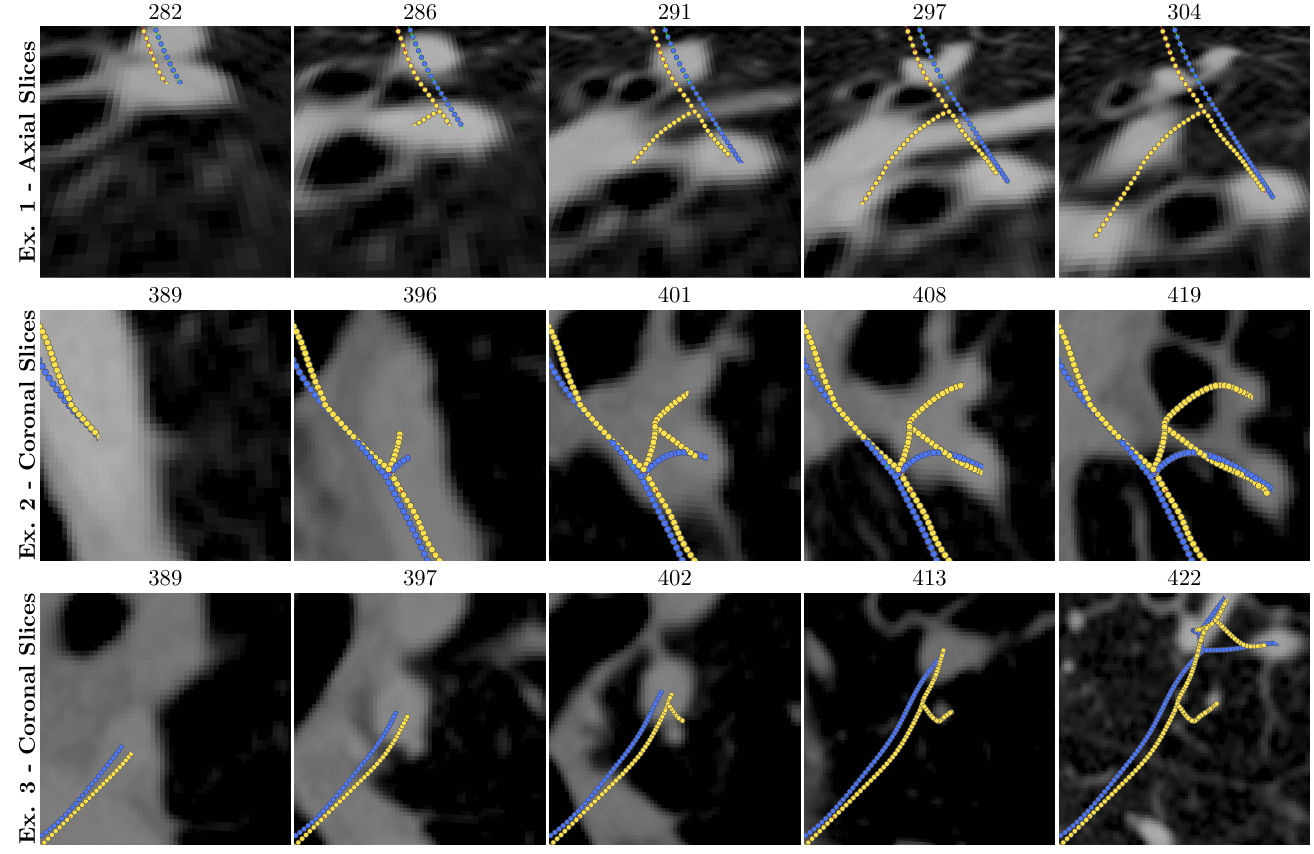}
\caption{Qualitative examples of failure cases where RefTr (blue) misses a vessel bifurcation that leads to a new ground truth vessel (yellow).}
\label{fig:bif_miss_pred}
\end{figure}

Figure~\ref{fig:bif_miss_pred} shows cases in which RefTr fails to predict a bifurcation and therefore misses a new branch.
In the first two examples, the bifurcating vessels remain partially connected across several slices before fully separating, making the bifurcation ambiguous.
In the third example, RefTr misses a relatively thin branch compared with the parent vessel.
These cases reduce recall at both the point and branch levels.

\begin{figure}[!ht]
\centering
\includegraphics[width=1.0\textwidth]{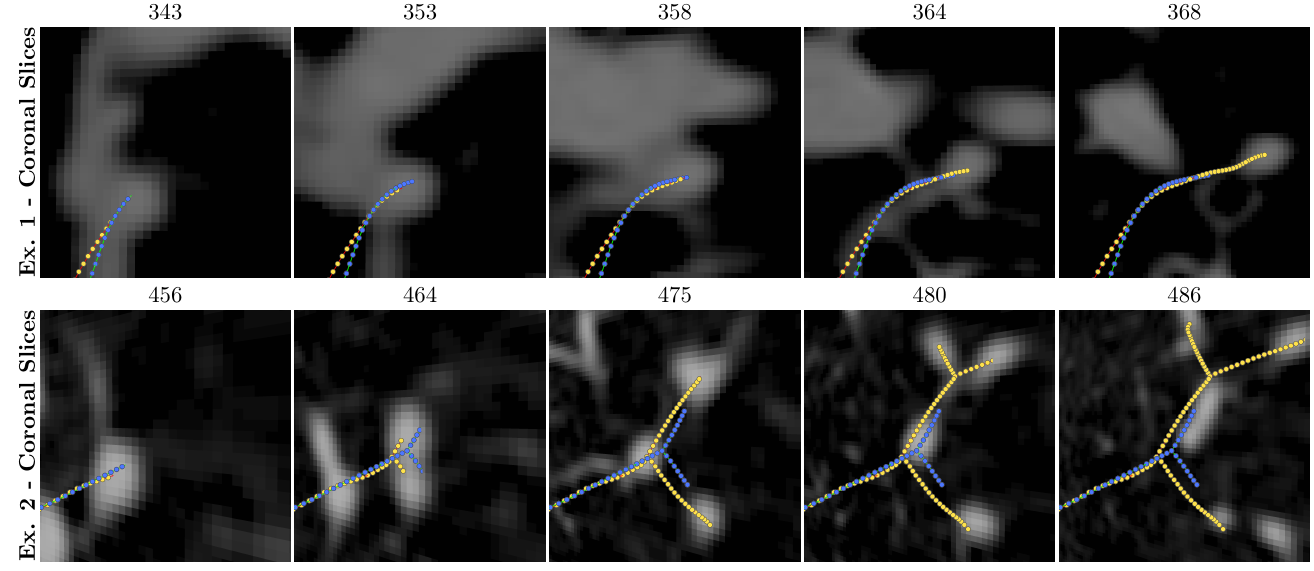}
\caption{Qualitative examples of failure cases where RefTr (blue) terminates a branch earlier than the ground truth (yellow).}
\label{fig:early_term_pred}
\end{figure}

Figure~\ref{fig:early_term_pred} illustrates another failure mode in which RefTr terminates a tracked branch prematurely.
In some cases, the model appears uncertain whether the branch continues past an ambiguous region or belongs to a different structure.
In other cases, the prediction appears to favor an earlier branch endpoint than the reference annotation.
These cases also reduce recall.

Such failures could potentially be mitigated by increasing the diversity and amount of training data, applying hard-mining strategies for difficult vessels, or introducing additional data augmentation.
We leave these directions to future work.

\end{document}